\documentclass{article}
\usepackage{spconf,amsmath,graphicx, amssymb, booktabs, makecell, multirow}
\usepackage{algorithm}
\usepackage{algorithmicx}
\usepackage{algpseudocode}
\usepackage{arydshln}
\usepackage{ifsym}
\newcommand{\ie}{\textit{i}.\textit{e}.}
\newcommand{\eg}{\textit{e}.\textit{g}.}

\title{Visually Dehallucinative Instruction Generation}
\name{Sungguk Cha, Jusung Lee, Younghyun Lee, Cheoljong Yang\sthanks{Corresponding Author}}
\address{Vision AI Lab., NC Research, NCSOFT Corporation}
\begin{document}
%\ninept
%
\maketitle
%
% 100 - 150 words
\begin{abstract}
% Following script contains 126 words
% Opening Statement/Context Establishment
In recent years, synthetic visual instructions by generative language model have demonstrated plausible text generation performance on the visual question-answering tasks. 
% Problem Statement/Objective
However, challenges persist in the hallucination of generative language models, \ie, the generated image-text data contains unintended contents.
% Methodology/Approach
This paper presents a novel and scalable method for generating visually dehallucinative instructions, dubbed \textit{CAP2QA},  that constrains the scope to only image contents.
% Key Contributions/Innovations
Our key contributions lie in introducing image-aligned instructive QA dataset \textit{CAP2QA-COCO} and its scalable recipe.
% Experimental Setup/Results
In our experiments, we compare synthetic visual instruction datasets that share the same source data by visual instruction tuning and conduct general visual recognition tasks. 
It shows that our proposed method significantly reduces visual hallucination while consistently improving visual recognition ability and expressiveness.
% Limitations/Future Work
% Concluding Remarks/Call to Action
\end{abstract}
\section{Introduction}
\label{sec:intro}
\begin{figure}[!ht]
\centering
\includegraphics[width=1.0\linewidth]{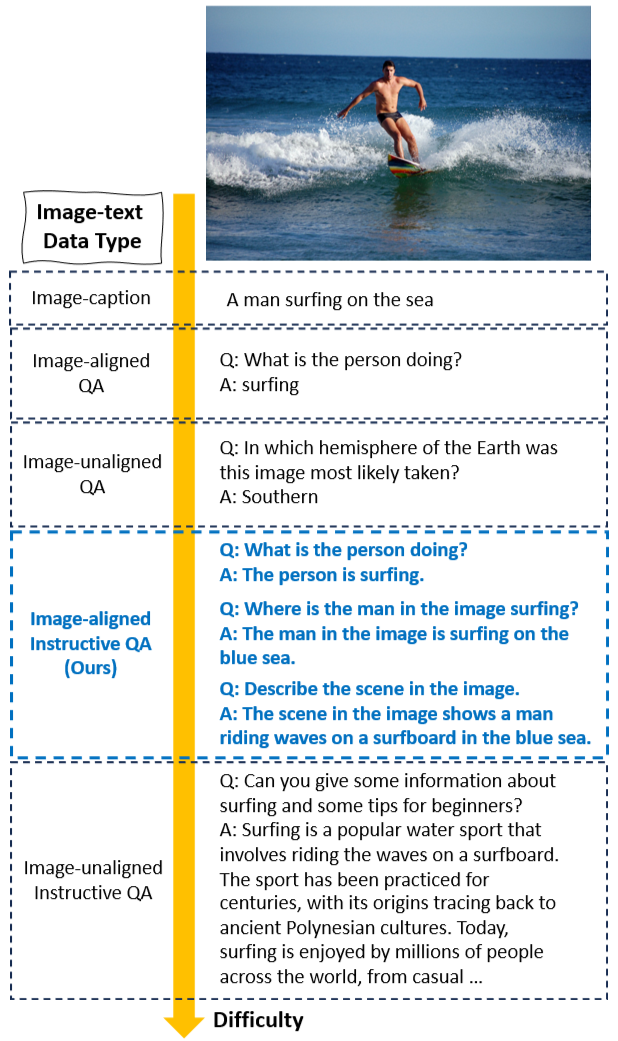}
\caption{
Types and samples of image-text data.
Image-alignness indicates if the context of QA is relevant to the image content.
Instructive QA refers to sentence-level-answer data.
Difficulty level regards visual question answering.
%Vision-language multi-modal data types with respect to the question contents (if it queries image-contents) and the answer formats (if it is a single word or sentence).
}
\label{fig:main}
\end{figure}
\begin{table*}[ht]
\centering
\caption{VQA datasets and their features.}
\resizebox{1.0\textwidth}{!}{
\begin{tabular}{ccccccccc}

    \toprule
    \multirow{2}{*}{Dataset} & Avg. \#word & \multirow{2}{*}{\#Image} & \multirow{2}{*}{\#Question} & \multirow{2}{*}{Scalable} & \multirow{2}{*}{ImageAligned} & \multirow{2}{*}{Recognition} & \multicolumn{2}{c}{Generation} \\
    & Question/Answer & & & & & & Description & Reasoning \\
    \midrule
    DAQUAR & 11.5/1.1 (word) & 1,449 & 12,468 & $\times$ & \checkmark & \checkmark & $\times$ & $\times$ \\
    VQAv2 & 6.1/1.2 (word) & 200k & 1.1M & $\times$ & \checkmark & \checkmark & $\times$ & $\times$ \\
    OKVQA & 8.1/1.3 (word) & 14,031 & 14,055 & $\times$ & $\times$ & \checkmark & $\times$ & \checkmark \\
    \hdashline[2pt/5pt]
    LLaVA & 10.7/\textbf{60.7} (sentence) & 80,000 & 221,333 & \checkmark & $\times$ & \checkmark & \checkmark & \checkmark \\
    \makecell{CAP2QA-COCO (Ours)} & 7.2/\textbf{5.4} (sentence) & 122,906 & 873,631 & \checkmark & \checkmark & \checkmark & \checkmark & \checkmark \\
    \bottomrule

\end{tabular}
}
\label{table:datasets}
\end{table*}

% Introduction to multimodal instruction data synthesis for VLM
Recent studies revealed the feasibility of visual instruction tuned vision-language model (VLM)~\cite{liu2023visual, dai2023instructblip, zhu2023minigpt} in addressing visual recognition tasks, including image captioning~\cite{chen2015microsoft} and visual question answering (VQA)~\cite{goyal2017making, marino2019ok}.
The central premise of VLM approach revolves around aligning visual embeddings with the semantic space of language, thereby facilitating comprehension and text generation by large language models (LLMs).
Upon achieving the alignment between vision and language, the VLM demonstrates enhanced zero-shot capabilities across novel tasks, necessitating the process of visual instruction tuning. 
Consequently, the availability of comprehensive vision-language datasets assumes a paramount role in this alignment process; however, the acquisition of such multi-modal data presents ongoing challenges.

% Initial works and their drawbacks
Previous studies~\cite{dai2023instructblip, zhu2023minigpt, gong2023multimodalgpt} have proposed repurposing existing datasets, such as those used in VQA and image-captioning tasks, to alleviate this scarcity.
Methods such as reformatting these datasets have been suggested, but these result in only one or two-worded answer available, which limits their use in descriptive and complex answering tasks.
To generate longer text, the approach~\cite{liu2023visual} to use GPT-assisted visual instruction generation has been brought up but has been criticized for being inaccurate and hallucinative~\cite{li2023seedbench, li2023evaluating}. %rohrbach2019object

%To address the limited availability of visual instruction data, recent studies~\cite{liu2023visual, dai2023instructblip, zhu2023minigpt} suggest repurposing existing image-text datasets, such as those used in VQA and image-captioning tasks.
%The InstructBLIP~\cite{dai2023instructblip} and MiniGPT-4~\cite{zhu2023minigpt} studies use methods of reformatting these datasets (for instance, transforming a VQA triple of an image-\{question\}-\{answer\} into an image-"Question: \{question\}. Short answer: \{answer\}").
%However, these transformations only seem to generate one or two-word phrases for VQA, as they utilize training datasets that include only single-word VQA solutions, which limits in descriptive tasks and complex answering tasks (Table~\ref{table:datasets}).
%To realize the generation of longer text, LLaVA~\cite{liu2023visual} brought forward a method involving GPT~\cite{openai2023gpt}-assisted visual instruction generation, which asks complex reasoning questions and allows for a sentence- or paragraph-level question-answering model.
%Nevertheless, quantitative analyses~\cite{li2023evaluating, li2023seedbench} critique the sentence-level approaches for being hallucinative and inaccurate.

% Contributions
%In this study, we introduce a novel and scalable method for generating visually dehallucinative instruction.
Our contributions are three-fold:
\begin{itemize}
  \item \textbf{Image-aligned instruction generation:} 
  We propose a scalable data synthesis method that generates visual instructions and ensures the image-text alignness of the generated text (Fig.~\ref{fig:main}). 
  This approach employs GPT-assistance with image-aligning prompts and filtering. 
  \item \textbf{Visually dehallucinative:} 
  Our experiments confirm the visual dehallucination on the object recognition tasks as well as the improvements in zero-shot visual recognition and expressiveness.
  \item \textbf{Large-scale multi-modal data:} 
  Lastly, we publicly release a large-scale multi-modal dataset, \textit{CAP2QA-COCO}, created using our proposed method with the COCO-caption dataset. \footnote{https://github.com/ncsoft/cap2qa}
\end{itemize}
\section{Related Work}
This section reviews the progression of visual question answering (VQA) datasets as seen in Fig.~\ref{fig:main} and Table~\ref{table:datasets}, leading to the subsequent need for generating visual instruction data.

\textbf{Evolution of VQA datasets}
VQA, the model-driven process of answering questions about an image in natural language, originally targeted generic recognition tasks involving object type, attribute, count, action, location, relative position, OCR, and more in datasets such as DAQUAR~\cite{malinowski2014multi} and VQAv2~\cite{goyal2017making} (image-aligned QA). 
OKVQA~\cite{marino2019ok} even demands external knowledge and reasoning to answer questions that fall outside the scope of image content, leading to a lack of image-alignment (image-unaligned QA).

\textbf{Transition from word-level to sentence-level answers}
As reflected in the 'Avg. \#word column' of Table~\ref{table:datasets}, VQA initially focused on one-word answers, barring few compound-word exceptions (e.g., 'new york'). 
In the pre-LLM era, text generation was a daunting task, transforming VQA into primarily a classification task. 
The advent of LLM signified a paradigm shift, making text generation viable. 
LLM-enabled models~\cite{liu2023visual, dai2023instructblip, zhu2023minigpt, gong2023multimodalgpt, ye2023mplugowl} exhibited the plausibility of sentence-level VQA, subsequently highlighting the need for visual instruction data. 
InstructBLIP~\cite{dai2023instructblip} reformatted image-text datasets into a "Q:\{the predefined instruction\}? A: \{answer\}" format, which resulted in artificially lengthy instructions, but retained concise, single-word answers.
On the other hand, LLaVA~\cite{liu2023visual} employed image annotations and GPT-assistance to generate complex reasoning QA that yielded sentence-level answers (image-unaligned instructive QA), which enabled VLM to generate detailed answers in sentence-level.
Presently, there is a lack of datasets designed specifically for the image-aligned instructive QA task. 

\section{CAP2QA}
Language-model-generated data has potential risks of hallucination. %~\cite{ji2023survey}.
Our motivation is to reduce the hallucination by allowing language models to generate only within the verified contents (\eg, human annotated image-caption data).

\begin{algorithm}
\caption{CAP2QA}
\label{algo:cap2qa}
\begin{algorithmic}[1]
    \Require \textit{GPT}: language model assistant, \textit{FilterArtifacts}: filtering function, \textit{prompt}, \textit{caption}, \textit{RETRY}: integer
    \For{$i = 1$ to RETRY}
        \State QAs $\gets$ GPT(prompt, caption)
        \State QAs $\gets$ FilterArtifacts(QAs)
        \If{len(QAs) $>$ 0}
            \State break
        \EndIf
    \EndFor
    \State \Return QAs
\end{algorithmic}
\end{algorithm}

% cap2qa method
We propose the vision-language-aligned question-answer generating framework, \textit{CAP2QA}, adopting text-only GPT (\textit{GPT-3.5-turbo} or \textit{GPT-4}) using image-aligned captions (Algorithm~\ref{algo:cap2qa}).
First, GPT generates QAs given the captions with detailed prompts composed of (1) bottom rules, (2) task description, and (3) conditions. 
The bottom rules emphasize to stick to the given context (\eg, "\textit{Any content should be included by the given context.}"), which enhances the image-alignment, and prevent from generating rhetoric texts in order to reduce post-process costs (\eg, "\textit{Must answer to what asked. In other word, if not asked, nothing should be answered such as explanation neither any comment.}").
The task description prompt prioritizes the generated question be answerable within the caption and limits the generated answer to contain excuses such as \textit{"Must avoid questioning and answering 'not specified in the caption'"}.
The condition prompts provide further details for inhibiting improper generation (\eg, "\textit{The generated question-answer should be reasonable and question should not imply answer.}" and "\textit{Must avoid question-answer if answering the question is not valid due to no specification}"). 
Next, it filters inappropriate artifacts for question-answering scenario such as mentioning \textit{'caption'} (\textit{"Question: Is the dessert 'in the caption' meant for one person or two people?, Answer: It is meant for two people 'according to the caption'"}).

\section{Experimental Results}
In this section, we compare CAP2QA-COCO with LLaVA, the unique sentence-level answering visual instruction data.
Note, the both datasets share the same source, MSCOCO. 

\subsection{Implementation Detail}
% CAP2QA COCO 뭐가 나왔는지 설명
\textbf{CAP2QA-COCO} 
We adopted ChatGPTs (\textit{GPT-3.5-turbo} and \textit{GPT-4})~\cite{openai2023gpt} for the language model assistant.
Using COCO-caption~\cite{chen2015microsoft} train and val set which has 591,753 and 25,014 captions respectively, CAP2QA generated 835,031 and 38,600 question-answer pairs. 
The generated instructions refer 117,933 and 4,973 images where the numbers of the source images are 118,287 and 5,000 for train and val set respectively. 

% visual instruction finetune 설명
\textbf{Visual instruction finetune}
For the fair comparison, we begin with the same pretrained VLM and visual-instruction-finetune them with LLaVA and CAP2QA-COCO respectively. 
We adopt the state-of-the-art VLM, InstructBLIP (ViT-G~\cite{dosovitskiy2021image} for the image encoder and OPT~\cite{zhang2022opt} for the language model) as a baseline.
We utilize various scaled language models %in VQA and image captioning experiments (Table~\ref{table:quantitative}) 
to explore the relationship between language model size and visual recognition performance, as well as for conducting ablation studies.
We perform visual instruction finetune the same pretrained InstructBLIP using LLaVA and CAP2QA-COCO for each with the same training protocol.
We implemented our method including InstructBLIP and brought the pretrained BLIP-2 from LAVIS framework~\cite{li-etal-2023-lavis}.
We employ AdamW optimizer with $\beta_1$ = 0.9 and $\beta_2$ = 0.999.
The learning rate ranges $10^{-6}$ to $10^{-4}$ for the warm up; then it decays to $10^{-5}$ with cosine scheduler. 
The models are trained for 5 epochs (10,000 steps per an epoch) with batch size 8 on 8 A100 GPUs.

\begin{table}[ht]
\centering
\caption{
Object hallucination evaluation on COCO validation set.
$CHAIR_s$ means answer-level hallucination rate.
$Recall$ refers object recognition recall.
$Recall_{w/oH}$ indicates recall excluding hallucinated answers. 
}
\vspace{0.2cm}
\resizebox{1.0\linewidth}{!}{
\begin{tabular}{c|ccc}
    \toprule
    Instruction Data & $CHAIR_s$ $(\downarrow)$ & $Recall$ $(\uparrow)$ & $Recall_{w/o H}$ $(\uparrow)$ \\
    \midrule
    LLaVA & 76\% & 0.21 & 0.07 \\
    CAP2QA-COCO & \textbf{9\%} & \textbf{0.32} & \textbf{0.3} \\
    \bottomrule
\end{tabular}
}
\label{table:hal}
\end{table}

\subsection{Evaluation}
We measure 
(1) the visual hallucination on COCO object recognition task~\cite{lin2015microsoft} with sentence-level object hallucination metric, $CHAIR_s$~\cite{rohrbach2019object};
(2) the visual recognition performance with VQA accuracy on VQAv2~\cite{goyal2017making}, GQA~\cite{hudson2019gqa} and OKVQA~\cite{marino2019ok};
and (3) the expressiveness of the model with BLEU and CIDEr on COCO-caption~\cite{chen2015microsoft}.

\textbf{Object hallucination}
$CHAIR_s$ is the object hallucination rate of sentences (\ie, $CHAIR_s$ 40\% means 40\% of the answers include hallucination.), proposed for sentence-level hallucination evaluation~\cite{li2023evaluating, rohrbach2019object}.
We conduct object recognition on COCO dataset, where we have positive and negative object category annotations per an image, by querying \textit{"Question: What objects are in the image?"}.
Next, we calculate $CHAIR_s$ by checking if each answer contains nonexistent sample.
We further provide recall with and without the hallucination, preventing anomaly cases such as saying nothing which has 0 hallucination rate and saying everything which causes recall 100\%.

\textbf{Visual recognition}
VQAv2 and GQA cover broad recognition tasks such as object type, attribute and relative positions. 
OKVQA contains outside-knowledge-requiring visual reasoning QAs.
%We prompted \textit{"Question: \{question\} Short Answer: \{answer\}"} and \textit{"Question: \{question\} Answer: \{answer\}"} for measuring VQA accuracy (Acc) and parsing VQA accuracy (PAcc, Algorithm~\ref{algo:pacc}) respectively as in \cite{dai2023instructblip}.
PAcc simply checks if the target answer is substring of the generated text and conducts the same VQA accuracy calculation (Algorithm~\ref{algo:pacc}).
It makes sentence compatibility (\eg, answer: \textit{"fish"}. prediction: \textit{"they 'fish' on the boat"} PAcc is \textit{True} while Acc is \textit{False}) and flexibility to the inflections (\eg, answer: \textit{"walk"}. prediction: \textit{"walking"} PAcc is \textit{True} but Acc is \textit{False}).

\begin{algorithm}
\caption{Parsing VQA accuracy}
\label{algo:pacc}
\begin{algorithmic}[1]
    \Require \textit{GroundTruths}: list of ground truths (words), \textit{Prediction}: prediction in sentence
    \State Initialize \textit{Match} $\gets 0$
    \For{each \textit{GT} in \textit{GroundTruths}}
        \If{\textit{GT} is substring of \textit{Prediction}}
            \State \textit{Match} += 1
        \EndIf
    \EndFor
    \State \Return min(1, \textit{Match} / 3)
\end{algorithmic}
\end{algorithm}
\vspace{-0.5cm}

\begin{table*}[ht]
\centering
\caption{Experimental results on visual question answering and image captioning tasks.}
\resizebox{1.0\textwidth}{!}{
\begin{tabular}{cccccccccc}
\toprule
\multirow{3}{*}{\makecell{LLM}} & \multirow{3}{*}{\makecell{Instruction\\Tuning Data}} & \multicolumn{6}{c}{Zero-shot VQAs} & \multicolumn{2}{c}{COCO-Caption} \\
& & \multicolumn{2}{c}{VQAv2-val} & \multicolumn{2}{c}{GQA} & \multicolumn{2}{c}{OKVQA} & \multirow{2}{*}{\makecell{BLEU-4 \small{($\uparrow$})}} & \multirow{2}{*}{\makecell{CIDEr \small{($\uparrow$})}} \\
& & Acc & PAcc & Acc & PAcc & Acc & PAcc & &\\
\midrule
\multirow{2}{*}{OPT-1.3B} & LLaVA & 0.0 & 0.496 & 0.0 & 0.35 & 0.0 & 0.185 & 0.068 & 0.012 \\
& CAP2QA-COCO & \textbf{0.177} & \textbf{0.568} & \textbf{0.062} & \textbf{0.404} & \textbf{0.074} & \textbf{0.361} & \textbf{0.209} & \textbf{0.827} \\ \hline%\cmidrule{2-11}
\multirow{2}{*}{OPT-2.7B} & LLaVA & 0.0 & 0.517 & 0.0 & 0.358 & 0.0 & 0.22 & 0.13 & 0.049 \\
& CAP2QA-COCO & \textbf{0.147} & \textbf{0.578} & \textbf{0.042} & \textbf{0.406} & \textbf{0.064} & \textbf{0.389} & \textbf{0.178} & \textbf{0.752} \\ \hline%\cmidrule{2-11} 
\multirow{2}{*}{OPT-6.7B} & LLaVA & 0.001 & 0.473 & 0.0 & 0.339 & 0.001 & 0.268 & 0.126 & 0.451 \\
& CAP2QA-COCO & \textbf{0.162} & \textbf{0.585} & \textbf{0.053} & \textbf{0.425} & \textbf{0.093} & \textbf{0.411} & \textbf{0.189} & \textbf{0.717} \\
\bottomrule
\end{tabular}
}
\label{table:quantitative}
\end{table*}

\subsection{Results}
\begin{figure}[!h]
\centering
\begin{minipage}[]{0.9\linewidth}
  \centering
  \centerline{\includegraphics[width=1.0\linewidth]{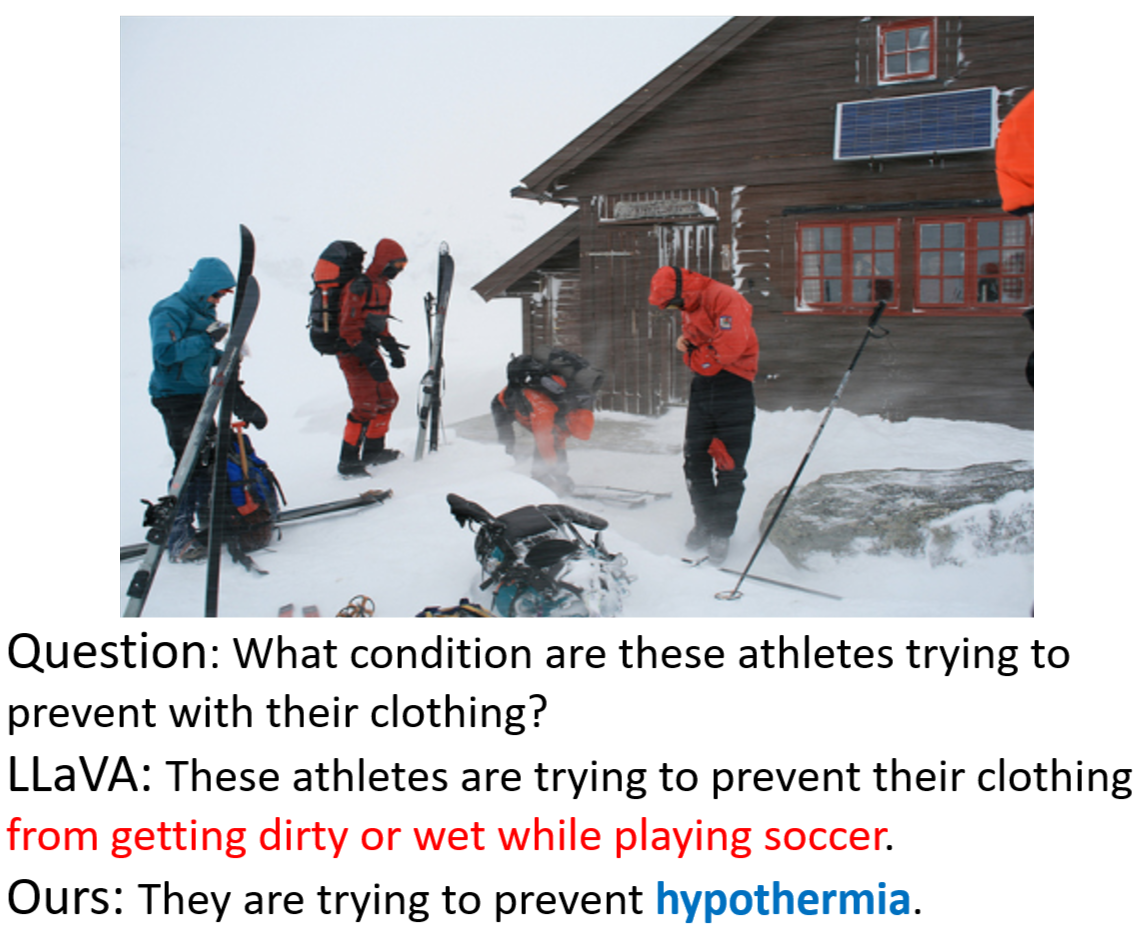}}
%  \vspace{2.0cm}
  \centerline{(a) Scene understanding }\medskip
\end{minipage}
\begin{minipage}[]{0.9\linewidth}
  \centering
  \centerline{\includegraphics[width=0.85\linewidth]{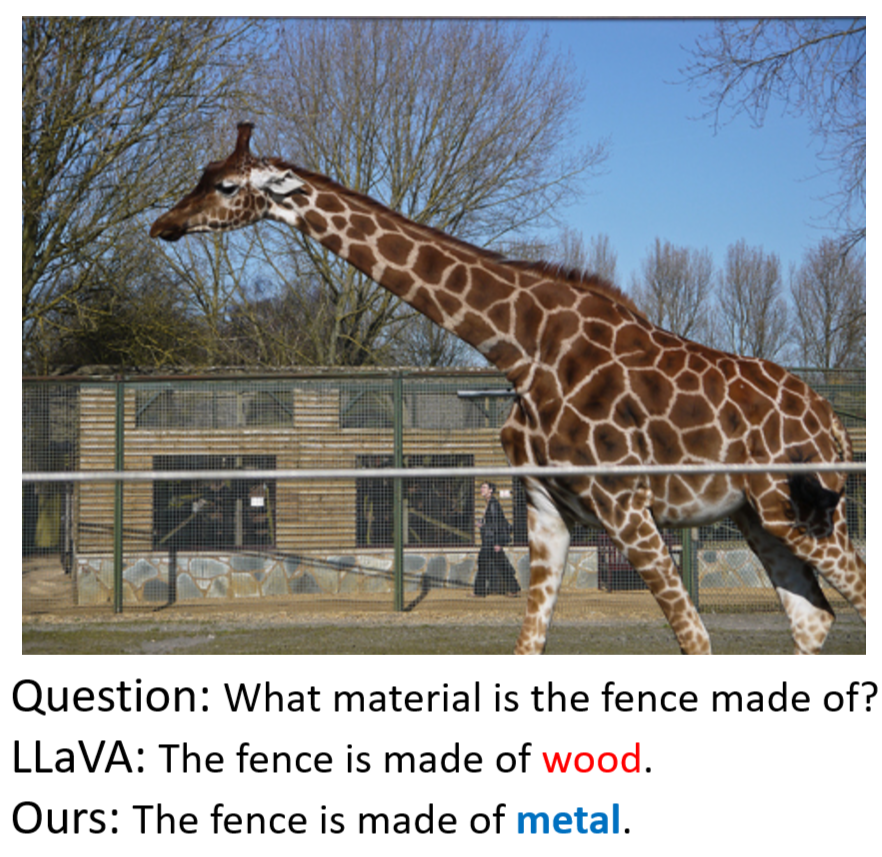}}
%  \vspace{1.5cm}
  \centerline{(b) Object understanding }\medskip
\end{minipage}
\caption{
Qualitative results in visual understanding. 
}
\label{fig:qual}
\vspace{-0.3cm}
\end{figure}

\textbf{Visual hallucination}
Table~\ref{table:hal} shows object hallucination evaluation on COCO validation set with InstructBLIP (ViT-G, OPT-1.3B).
CAP2QA-COCO achieves remarkably less hallucination rate ($CHAIR_s$ $9\%$ vs $76\%$) and recognition performance in $Recall$ ($32\%$ vs $21\%$).
In particular, $Recall_{w/oH}$ demonstrates CAP2QA-COCO tuned model results more accurate even without visual hallucination.
Fig.~\ref{fig:qual} illustrates visual hallucination cases.
The red texts in the figure are irrelevant to the image or incorrect, while ours in the blue texts responses appropriately. 
%On the other hand, the image-unaligned dataset shows severe hallucination in both $CHAIR_s$ and the recall drop $Recall - Recall_{w/oH}$.

\textbf{Visual question answering}
Zero-shot VQA metrics depicted in Table~\ref{table:quantitative} reflect the generalized visual recognition performance over objects, attribute, location relative positions and more. 
The absence of negative annotation makes hallucination evaluation invalid, thus only VQA accuracy (similar to recall) is reported.
CAP2QA-COCO significantly outperformed on the general benchmarks (the worst CAP2QA-COCO in 1.3B outperforms the best LLaVA in 6.7B).
The consistent increments in PAcc with respect to language model size regardless of the instruction data imply larger (or better) language model leads to better visual recognition performance (VQAv2 and GQA) and complex reasoning (OKVQA). 

\textbf{Image captioning}
The caption metrics, so called expressiveness metrics, compare the generated text with the original caption data (\ie, the higher score, the more similar semantically).
Our experiment suggests CAP2QA-COCO preserves the source knowledge more and LLaVA tends to generate contents irrelevant to the source.
Ours shows slightly decreasing scores over the language model sizes, indicating the smaller language model, the more fit to the tuning data.
On the other hand, considering the increasing tendency with respect to the model scale of LLaVA tuned models, LLaVA seems too challenging for the visual instruction tuning. 

\section{Conclusion}
In this work, we present visually dehallucinative instruction generation, \textit{CAP2QA}.
We experimented sentence-level instruction generation methods with the same data source on the well known multi-modal visual recognition tasks. 
Our experiments proved the effectiveness of image-alignness in visual instruction generation, showing noteworthy advances in reducing visual hallucination; and consistent enhancements in zero-shot VQA and expressiveness.
In future research, it would be valuable to delve deeper into utilizing web-scale datasets which requires careful text data selection considering caption noise and the purpose of usages.

%In summary, image-alignness in visual instruction data alleviates visual hallucination. 

%\textbf{Future Work: Scale to Webscale}
%The COCO dataset is employed in our experimental work, which despite its high-quality human annotation, does have certain limitations regarding the extent of knowledge it encompasses.
%Only a select number of categories are addressed in the COCO dataset not only causing a dearth of diversity in image categories but similarly, the text data obtained is also considerably scant.
%To cover up broad knowledge, future work should explore methodologies for extracting or filtering data from existing webscale datasets, such as LAION~\cite{schuhmann2021laion400m}.
%SBU~\cite{NIPS2011_5dd9db5e}, Conceptual Captions~\cite{sharma2018conceptual} and .
% CC12M changpinyo2021conceptual

\vfill\pagebreak

\newpage
\label{sec:refs}
\bibliographystyle{IEEEbib}
\bibliography{strings,refs}

\begin{thebibliography}{10}

\bibitem{liu2023visual}
Haotian Liu, Chunyuan Li, Qingyang Wu, and Yong~Jae Lee,
\newblock ``Visual instruction tuning,''
\newblock {\em arXiv preprint arXiv:2304.08485}, 2023.

\bibitem{dai2023instructblip}
Wenliang Dai, Junnan Li, Dongxu Li, Anthony Meng~Huat Tiong, Junqi Zhao, Weisheng Wang, Boyang Li, Pascale Fung, and Steven Hoi,
\newblock ``Instructblip: Towards general-purpose vision-language models with instruction tuning,''
\newblock {\em arXiv preprint arxiv:2305.06500}, 2023.

\bibitem{zhu2023minigpt}
Deyao Zhu, Jun Chen, Xiaoqian Shen, Xiang Li, and Mohamed Elhoseiny,
\newblock ``Minigpt-4: Enhancing vision-language understanding with advanced large language models,''
\newblock {\em arXiv preprint arXiv:2304.10592}, 2023.

\bibitem{chen2015microsoft}
Xinlei Chen, Hao Fang, Tsung-Yi Lin, Ramakrishna Vedantam, Saurabh Gupta, Piotr Doll{'a}r, and C~Lawrence Zitnick,
\newblock ``Microsoft coco captions: Data collection and evaluation server,''
\newblock in {\em Proceedings of the IEEE International Conference on Computer Vision}, 2015, p. 3570–3577.

\bibitem{goyal2017making}
Yash Goyal, Tejas Khot, Douglas Summers-Stay, Dhruv Batra, and Devi Parikh,
\newblock ``Making the v in vqa matter: Elevating the role of image understanding in visual question answering,''
\newblock in {\em Proceedings of the IEEE Conference on Computer Vision and Pattern Recognition}, 2017, pp. 6904--6913.

\bibitem{marino2019ok}
Kenneth Marino, Mohammad Rastegari, Ali Farhadi, and Roozbeh Mottaghi,
\newblock ``Ok-vqa: A visual question answering benchmark requiring external knowledge,''
\newblock in {\em Proceedings of the IEEE Conference on Computer Vision and Pattern Recognition}, 2019, pp. 3195--3204.

\bibitem{gong2023multimodalgpt}
Tao Gong, Chengqi Lyu, Shilong Zhang, Yudong Wang, Miao Zheng, Qian Zhao, Kuikun Liu, Wenwei Zhang, Ping Luo, and Kai Chen,
\newblock ``Multimodal-gpt: A vision and language model for dialogue with humans,''
\newblock {\em arXiv preprint arXiv:2305.04790}, 2023.

\bibitem{li2023seedbench}
Bohao Li, Rui Wang, Guangzhi Wang, Yuying Ge, Yixiao Ge, and Ying Shan,
\newblock ``Seed-bench: Benchmarking multimodal llms with generative comprehension,''
\newblock {\em arXiv preprint arXiv:2307.16125}, 2023.

\bibitem{li2023evaluating}
Yifan Li, Yifan Du, Kun Zhou, Jinpeng Wang, Wayne~Xin Zhao, and Ji-Rong Wen,
\newblock ``Evaluating object hallucination in large vision-language models,''
\newblock {\em arXiv preprint arXiv:2305.10355}, 2023.

\bibitem{malinowski2014multi}
Mateusz Malinowski and Mario Fritz,
\newblock ``A multi-world approach to question answering about real-world scenes based on uncertain input,''
\newblock in {\em Advances in Neural Information Processing Systems}, 2014, vol.~27.

\bibitem{ye2023mplugowl}
Qinghao Ye, Haiyang Xu, Guohai Xu, Jiabo Ye, Ming Yan, Yiyang Zhou, Junyang Wang, Anwen Hu, Pengcheng Shi, Yaya Shi, Chenliang Li, Yuanhong Xu, Hehong Chen, Junfeng Tian, Qian Qi, Ji~Zhang, and Fei Huang,
\newblock ``mplug-owl: Modularization empowers large language models with multimodality,''
\newblock {\em arXiv preprint arXiv:2304.14178}, 2023.

\bibitem{openai2023gpt}
R~OpenAI,
\newblock ``Gpt-4 technical report,''
\newblock {\em arXiv}, pp. 2303--08774, 2023.

\bibitem{dosovitskiy2021image}
Alexander Kolesnikov, Alexey Dosovitskiy, Dirk Weissenborn, Georg Heigold, Jakob Uszkoreit, Lucas Beyer, Matthias Minderer, Mostafa Dehghani, Neil Houlsby, Sylvain Gelly, Thomas Unterthiner, and Xiaohua Zhai,
\newblock ``An image is worth 16x16 words: Transformers for image recognition at scale,''
\newblock in {\em International Conference on Learning Representations}, 2021.

\bibitem{zhang2022opt}
Susan Zhang, Stephen Roller, Naman Goyal, Mikel Artetxe, Moya Chen, Shuohui Chen, Christopher Dewan, Mona Diab, Xian Li, Xi~Victoria Lin, Todor Mihaylov, Myle Ott, Sam Shleifer, Kurt Shuster, Daniel Simig, Punit~Singh Koura, Anjali Sridhar, Tianlu Wang, and Luke Zettlemoyer,
\newblock ``Opt: Open pre-trained transformer language models,''
\newblock {\em arXiv preprint arXiv:2205.01068}, 2022.

\bibitem{li-etal-2023-lavis}
Dongxu Li, Junnan Li, Hung Le, Guangsen Wang, Silvio Savarese, and Steven~C.H. Hoi,
\newblock ``{LAVIS}: A one-stop library for language-vision intelligence,''
\newblock in {\em Proceedings of the 61st Annual Meeting of the Association for Computational Linguistics (Volume 3: System Demonstrations)}, Toronto, Canada, July 2023, pp. 31--41, Association for Computational Linguistics.

\bibitem{lin2015microsoft}
Tsung-Yi Lin, Michael Maire, Serge Belongie, James Hays, Pietro Perona, Deva Ramanan, Piotr Dollar, and Larry Zitnick,
\newblock ``Microsoft coco: Common objects in context,''
\newblock in {\em European Conference on Computer Vision}, September 2014.

\bibitem{rohrbach2019object}
Anna Rohrbach, Lisa~Anne Hendricks, Kaylee Burns, Trevor Darrell, and Kate Saenko,
\newblock ``Object hallucination in image captioning,''
\newblock in {\em Proceedings of the 2018 Conference on Empirical Methods in Natural Language Processing}.

\bibitem{hudson2019gqa}
Drew~A. Hudson and Christopher~D. Manning,
\newblock ``Gqa: A new dataset for real-world visual reasoning and compositional question answering,''
\newblock in {\em Proceedings of the IEEE Conference on Computer Vision and Pattern Recognition}, June 2019.

\end{thebibliography}

\end{document}